\documentclass{ifacconf}

\usepackage{tikz}
\usepackage{pgfplots}
\tikzstyle{block} = [draw, rectangle, 
    minimum height=3em, minimum width=5em]
\tikzstyle{input} = [coordinate]
\tikzstyle{output} = [coordinate]
\tikzstyle{pinstyle} = [pin edge={to-,thin,black}]
\usetikzlibrary{shapes.geometric,shapes.arrows,decorations.pathmorphing,calc}
\usetikzlibrary{matrix,chains,scopes,positioning,arrows,fit}
\usepackage{subfig}
\usepackage{dsfont}
\usepackage{multirow}
\usepackage{algorithm}
\usepackage[noend]{algcompatible}

\usepackage{amsmath}
\usepackage{steinmetz}
\usepackage{graphicx}      
\usepackage{natbib}        
\usepackage{amsfonts}
\pgfplotsset{compat=newest, ticks=none}
\begin{document}
\begin{frontmatter}

\title{Control-Tutored Reinforcement Learning: an application to the Herding Problem}


\author[First]{Francesco De Lellis}
\author[Second]{Fabrizia Auletta}
\author[Third]{Giovanni Russo}
\author[Fourth]{Mario di Bernardo}

\address[First]{Department of Electrical Engineering and Information Technology, University of Naples Federico II, Naples, Italy (e-mail: f.delellis@studenti.unina.it).}
\address[Second]{Department of Engineering Mathematics, University of Bristol, Bristol, U.K. (e-mail: fabrizia.auletta@bristol.ac.uk).}
\address[Third]{School of Electrical and Electronic Engineering, University College Dublin, Dublin 4, Ireland (e-mail: giovanni.russo1@ucd.ie)}
\address[Fourth]{Department of Electrical Engineering and Information Technology, University of Naples Federico II, Naples, Italy,\\Department of Engineering Mathematics, University of Bristol, Bristol, U.K.  (e-mail: mario.dibernardo@unina.it),}

\end{frontmatter}

\section{Extended abstract}
Model-free reinforcement learning (or simply reinforcement learning, RL, in what follows) is increasingly used in applications to solve a wide variety of control problems \citep{kober2013reinforcement, garcia2015comprehensive, cheng2019end}. The lack of requiring a formal model of the plant renders it appealing for a heuristic, low-cost control design approach that can be easily implemented and adapted to different situations. 
As a trade-off, learning processes often require a long training phase where the controller agent learns by trial-and-error how the plant responds to different control actions, and what actions to take to steer its behavior in a desired manner. This problem is particularly relevant when using tabular methods, such as Q-learning, in those situations where reinforcement learning is applied to control dynamical systems defined in continuous spaces \citep{lillicrap2015continuous}. 
It is therefore desirable to enhance the learning process by encoding some qualitative knowledge of the system dynamics via appropriate models. This is the idea of model-based reinforcement learning approaches which are becoming increasingly popular in the control community \citep{ Atkeson97, Kurutach18, kurutach2018model, FerraroRusso2019}. 
These strategies aim at incorporating into the learning process some model of the uncertain dynamics of the plant to achieve better performance and considerably decrease the learning times.

In this extended abstract we develop a novel {\em control-tutored} Q-learning approach (CTQL) as part of the ongoing effort in developing model-based and safe RL for continuous state spaces. Differently from the existing approaches in model-based RL e.g. \citep{ gu2016continuous, deisenroth2011pilco}, we suggest here that the use of a feedback control strategy, with very limited knowledge of the plant dynamics, can be effectively used to improve convergence of the learning process towards achieving the control goal.
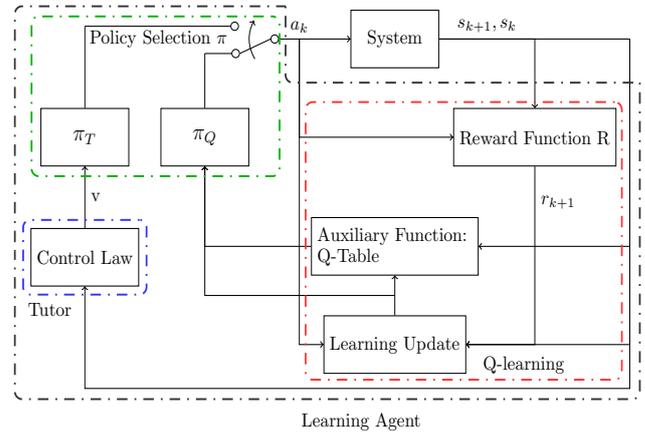
\begin{figure}[htbp]
\centering
\resizebox{8.4cm}{6cm}{
\begin{tikzpicture}[auto, node distance=1.8cm]
    
    \node [input, name=input] {};
    \node [block, right of=input] (system) {System};
    \node [block, below of=system, right of=system, xshift=1cm] (RF) {Reward Function R};
    \node [block, below of=system, yshift=-2cm, text width=3.1cm] (Q) {Auxiliary Function: Q-Table};
    \node [block, below of=Q] (LU) {Learning Update};
    \node [block, below of=system, left of=system, xshift=-2cm] (policyQ) {\large $\pi_Q$};
    \node [block, below of=system, left of=system, xshift=-4.4cm] (policyT) {\large $\pi_T$};
    \node [block, below of=policyT, yshift=-0.4cm] (CL) {Control Law};

    \draw [->] (system) -| node [pos=0.25] [name=S] {$s_{k+1},s_{k}$} (RF);
    \draw [-] (system) -- ++(4.7cm,0) coordinate(S1) {};
    \draw [->] (RF) |-  node [pos=0.1] {$r_{k+1}$} (LU);
    \draw [->] (S1) |- (Q);
    \draw [->] (Q) -| (policyQ);
    \draw [<-o] (system.west) -- node [pos=0.64] [name=A] [above] {$a_{k}$} ++(-1.6cm,0) coordinate(p1){};
    \draw [->] (A) |- (RF);
    \draw [->] (S1) |- (LU);
    \draw [->] (A) |- (LU);
    \draw [->] (LU) -- node [pos=0.5] [name=temp] [right] {} (Q);
    \draw [->] (temp) -| (policyQ);
    \draw [->] (S1) |- ([xshift=-0.3cm,yshift=-0.27cm]LU.south west) -| (CL);
    \draw [->] (CL) -- node [pos=0.5] [name=v] [right] {v} (policyT);
    \draw [-] (policyQ.north) -- ++(0,1cm) coordinate(pq1){};
    \draw [-o] (pq1) -- ++(0.7cm,0) coordinate(pq2){};
    \draw [-] (policyT.north) -- ++(0,1.5cm) coordinate(pt1){};
    \draw [-o] (pt1) -- ++(3.08cm,0) coordinate(pt2){};
    \draw [-] (p1) -- (pq2);
    \draw[<-] ($(pt2)+(0.3cm,0.1cm)$) to [bend right]($(pq2)+(0.3cm,-0.1cm)$);
    \tikzset{black dotted/.style={draw=black!80!white, line width=1pt, dash pattern=on 1pt off 4pt on 6pt off 4pt, inner sep=5mm, rounded corners}};
    \tikzset{blue dotted/.style={draw=blue!80!white, line width=1pt, dash pattern=on 1pt off 4pt on 6pt off 4pt, inner sep=1.5mm, rectangle, rounded corners}};
    \tikzset{red dotted/.style={draw=red!80!white, line width=1pt, dash pattern=on 1pt off 4pt on 6pt off 4pt, inner sep=1.1mm, rounded corners}};
    \tikzset{green dotted/.style={draw=green!70!black, line width=1pt, dash pattern=on 1pt off 4pt on 6pt off 4pt, inner sep=1.8mm, rounded corners}};

    \draw[-] [black dotted] ($(RF)+(2.1cm,1cm)$) |- ($(LU)+(0.1cm,-1cm)$);
    \draw[-] [black dotted] ($(LU)+(0.1cm,-1cm)$) -| node [pos=0.05, yshift=0.25cm] [name=LA] {Learning Agent} ($(CL)+(-1.4cm,0.1cm)$);
    \draw[-] [black dotted] ($(CL)+(-1.4cm,0.1cm)$) -- ($(policyT)+(-1.4cm,0.2cm)$) |- ($(p1)+(0cm,0.6cm)$) -- ++(0.3cm,0cm) coordinate(x1) {};
    \draw[-] [black dotted] (x1) |- ($(RF)+(2.1cm,1cm)$) {}; 
    
    \node (second dotted box) [blue dotted, fit = (CL)] {};
    \node at (second dotted box.south) [below, xshift=-0.7cm]  {Tutor};
    
    \node (third dotted box) [red dotted, fit = (Q) (LU) (RF)] {};
    \node at (third dotted box.south) [above, xshift=1.2cm]  {Q-learning};
    
    \node (fourth dotted box) [green dotted, fit = (policyT) (policyT) (p1) (pt2)] {};
    \node at (fourth dotted box.south) [above, xshift=0.06cm, yshift=2.25cm]  {Policy Selection $\pi$};
\end{tikzpicture}
}
\caption{Control Tutored Q-learning (CTQL) Schematic} 
\label{fig:CTQL}
\end{figure}

The key idea behind CTQL is schematically summarized  in Fig.~\ref{fig:CTQL}. 
Specifically, CTQL adopts the same Q-table structure and learning update rule  of classical Q-learning \citep{Sutton1998} but exploits a new policy selection function, say $\pi$.
As shown in Fig.~\ref{fig:CTQL}, at step $k$ the learning agent selects its next action $a_k$ in the action space $\mathcal{A}$ from a given system state $s_k$ belonging to the state space $\mathcal{S}$,
by choosing either the action suggested by the control tutor via the policy $\pi_T$ or the one suggested by the classical $\varepsilon$-greedy policy $\pi_Q$ used for the Q-learning. 

Mathematically, the policy selection function in CTQL is a switching policy defined as:
\begin{equation} \label{eq:4-6}
    \pi(a\vert s) = 
    \begin{cases} 
        \pi_Q(a\vert s),   &    \max_{a \in \mathcal{A}}\{Q(s,a)\}>0\\
        \pi_T(a),   &   \text{otherwise}
    \end{cases}
\end{equation}

In practice, before selecting its action, the agent checks the values stored in the Q-table for all the actions $a \in \mathcal{A}$.
If none of them has a positive value (i.e. the experience gained so far is not enough to decide what action to take), the learning agent follows the suggested action coming from the $\varepsilon$-greedy control tutor policy $\pi_T$, defined as follows:
\begin{equation} \label{eq:4-5}
  \pi_T(a) = 
    \begin{cases} 
        arg \min_{ a \in \mathcal{A}}\{\|v - a\|\}, &   \text{with probability ($1 -\varepsilon$)}\\
        rand(a),   &   \text{with probability $\varepsilon$}
    \end{cases}
\end{equation}
where $v$ is an action decided by a feedback control law with limited knowledge of the system dynamics and $\varepsilon$ is a positive constant in the range $]0,1[$ representing the probability of taking a random action, to promote exploration of the action space. 

The action $a_k$ selected by the switching policy $\pi$ is then used to update the Q-table.

We validate our approach by applying it to a challenging multi-agent {\em herding} control problem where $N$ mobile target agents in the plane must be collected and driven to a target goal region $G$ by $M$ herding agents \citep{Licitra2017,pierson2017controlling}. 

We assume the agents to be able to adjust their velocities almost instantaneously, as done for example in \cite{albi2016invisible}.

Assuming, the target agents' velocity is upper bounded by some maximum velocity $v_{t,max}$, the dynamics of the target agents is given by: 
\begin{equation} \label{eq:3-3}
    \dot{ x}^i_t =
    \begin{cases}
         f^i(x_t^i, x_h,t),     &    \|f^i(x_t^i, x_h,t)\| < v_{t,max}\\
        v_{t,max}e^{\jmath \scalebox{0.8}{\phase{f^i(x_t^i, x_h,t)}}}, &   \text{otherwise}
    \end{cases}
\end{equation}
where $ x_t^i \in \mathds{R}^2$ is the position of the $i$-th target agent, $x_h = [x_h^1, ..., x_h^M]$ is the vector of the positions in the plane of all $M$ herder agents, and the vector field $f^i:\mathds{R}^{2(M+1)}\times \mathds{R}\mapsto\mathds{R}^2$ is assumed to be the sum of two contributions, i.e. $f^i=f_1^{i}+f_2^i$.

Here, the term $f_1^{i}$ models the action of the herders onto the target and is defined as:
\begin{equation}
     f_1^i(x_t^i, x_h,t)  := \mu \sum_{j=1}^M \frac{ x_t^i -  x_h^j}{\| x_t^i -  x_h^j\|^3} U ( x_t^i,  x_h^j, \rho_t)
\end{equation} 
where $\mu$ is a constant gain modelling the intensity of the coupling with the herder and $U$ is an interaction function defined as:
\begin{equation} \label{eq:step}
    U ( x_t^i,  x_h^j, \rho_t) = \begin{cases}
    1, & \| x_t^i -  x_h^j\| < \rho_t \\ 
    0, & \text{otherwise} 
    \end{cases}
\end{equation}
that ensures that the coupling between target and herder agents is active only if their relative position is smaller than some target’s influence radius $\rho_t$.

The term $f_2^i$  represents the target own random dynamics defined as:
\begin{equation} 
      f_2^i(t) := \beta^i(t) e^{\jmath \theta^i(t)}
\end{equation}
where $\beta^i(t)$ and $\theta^i(t)$ are scalars updated every $\Delta t$ seconds from the uniform distributions $ \mathcal{U}(0,\beta_{max}) $ and $\mathcal{U}(0,2\pi)$, respectively.

The generic herder dynamics is given instead by:
\begin{equation} \label{eq:3-1}
  \dot{ x}_h^j(t) =   u^j(t)    \qquad \forall j = 1,...,M 
\end{equation}
where $u^j$ is a control input. 

The control objective is to design the input vector $  u = [  u^1,..., {u^M}]$ able to force targets to reach and remain in the circular goal region $ G := \{  x\in {\mathbb{R}}^2 : \| x -  x_g\|<\rho_g\}$  of center $ x_g$  and radius $\rho_g$. That is, to achieve the goal:
\begin{equation}
    \limsup_{t\to\infty} \| x_t^i(t) -  x_g\| < \rho_g \qquad \forall i = 1,...,N 
\end{equation}

{\bf Control Design.} For the sake of simplicity we consider the case where $N=M=1$ and the goal region is centered at the origin, i.e. $x_g=0$. We suppose the herder knows the position of the target but possesses only a conservative estimate, $\hat \rho_t<\rho_t$ of the target's true influence radius $\rho_t$.

We then design the control input $u$ driving the herder as follows. If $\|x_t-x_h\|>\hat \rho_t$ then the herder moves towards the target at its maximum speed in order to reduce its distance until entering the estimated influence region where $\|x_t-x_h\|\leq\hat \rho_t$. From this point, CTQL is used to drive the interaction between the herder and the target according to the following implementation (see \cite{DeLellis2019} for further details):

\begin{itemize}
\item The state space $\mathcal S$ is defined, for each herder, in terms of the possible discretized values of the relative distance between the herder and the target chased by it, the angular position of the herder, and the speed of the chased target. 

\item The action space $\mathcal A$ is chosen to be the set of possible discretized values of the input vector $ u^j$ to the herder dynamics given by \eqref{eq:3-1}.

\item The reward function $R$ (see Fig. \ref{fig:CTQL}) is selected as the sum of three contributions; one term that evaluates how close to the goal region the action takes the chased target to, one term that minimizes the distance between the herder and the chased target, and a final term that penalizes the herder if it enters the goal region after the action is taken. 

\item The tutoring control law $v$ in \eqref{eq:4-5} is computed assuming the herders only possess a rough model of the true target dynamics given by
\begin{equation}\label{eq:4-3}
    \dot{ x}_t = \delta( x_t -  x_h) U( x_t, x_h,\hat \rho_{t})
\end{equation} 
where $\delta$ stands for the intensity of the coupling with the herder, $U(\cdot)$ is the step function defined in~\eqref{eq:step}.
Specifically, using the following Lyapunov candidate function:
\begin{gather} \label{eq:4-7}
    V = \frac{1}{2} x_t^T x_t
\end{gather}
we choose the control law:
\begin{equation} \label{eq:4-9}
     v({ x}_t, \dot{ x}_t) = k\dot{ x}_t,  \qquad  k>1
\end{equation}
that guarantees:
\begin{gather} \label{eq:4-8}
    \dot{V} =  x_t^T\dot{ x}_t =  x_t^T( x_t -  x_h) <0
\end{gather}
so that the relative distance between the herder and the center of the goal region decays to zero if \eqref{eq:4-3} were a good model of the target dynamics.
\end{itemize}

\begin{figure}[htbp]
\includegraphics[width=\linewidth]{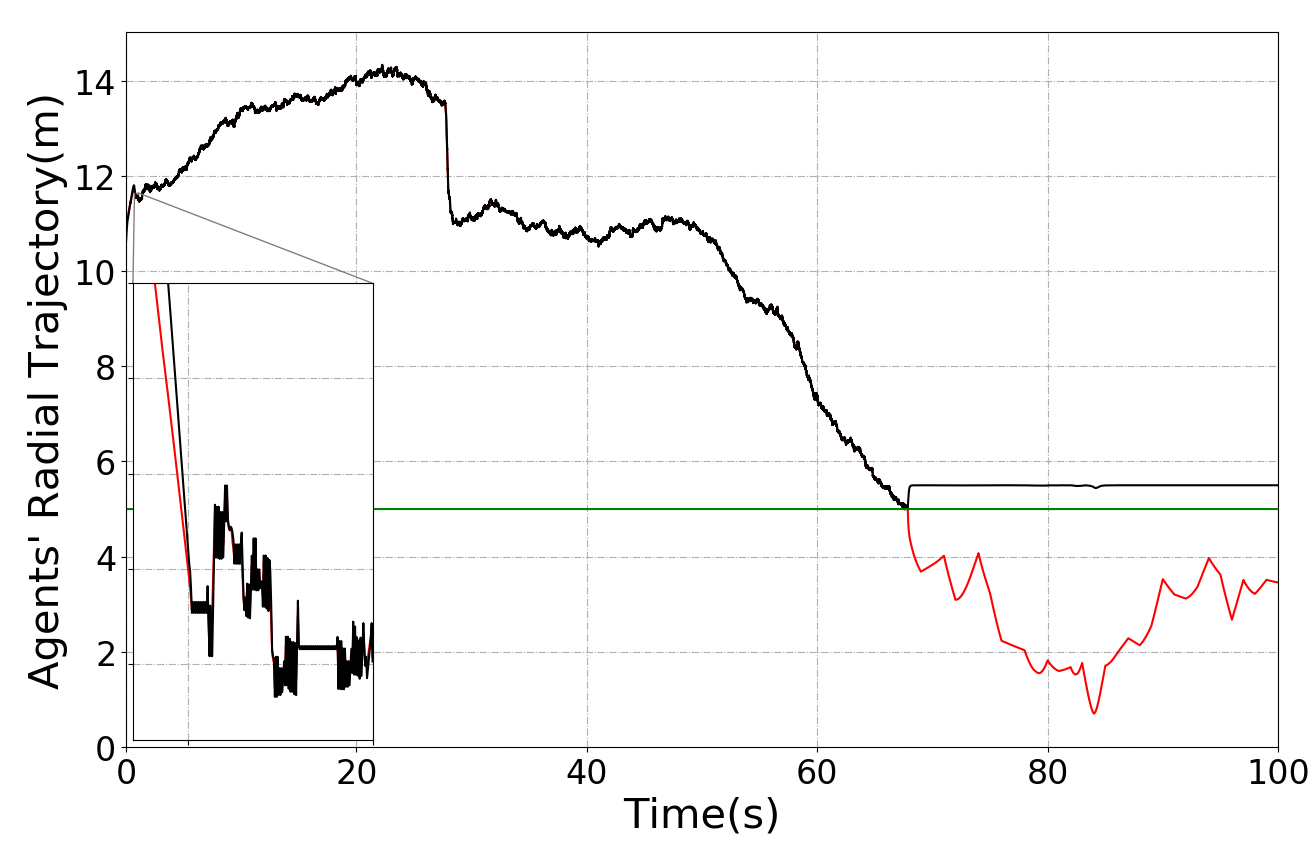}    
\caption{Radial coordinate of the target (red line) and of the herder (black line) driven by the CTQL algorithm after just one trial. The green solid line shows the radius of the circular region. The inset highlights a zoom of the transient dynamics during the interval $t \in [0.74, 1]s$.}
\label{fig:ATTVFB200}
\end{figure}

\begin{figure}[htbp]
\subfloat[ ]{\includegraphics[width= \linewidth]{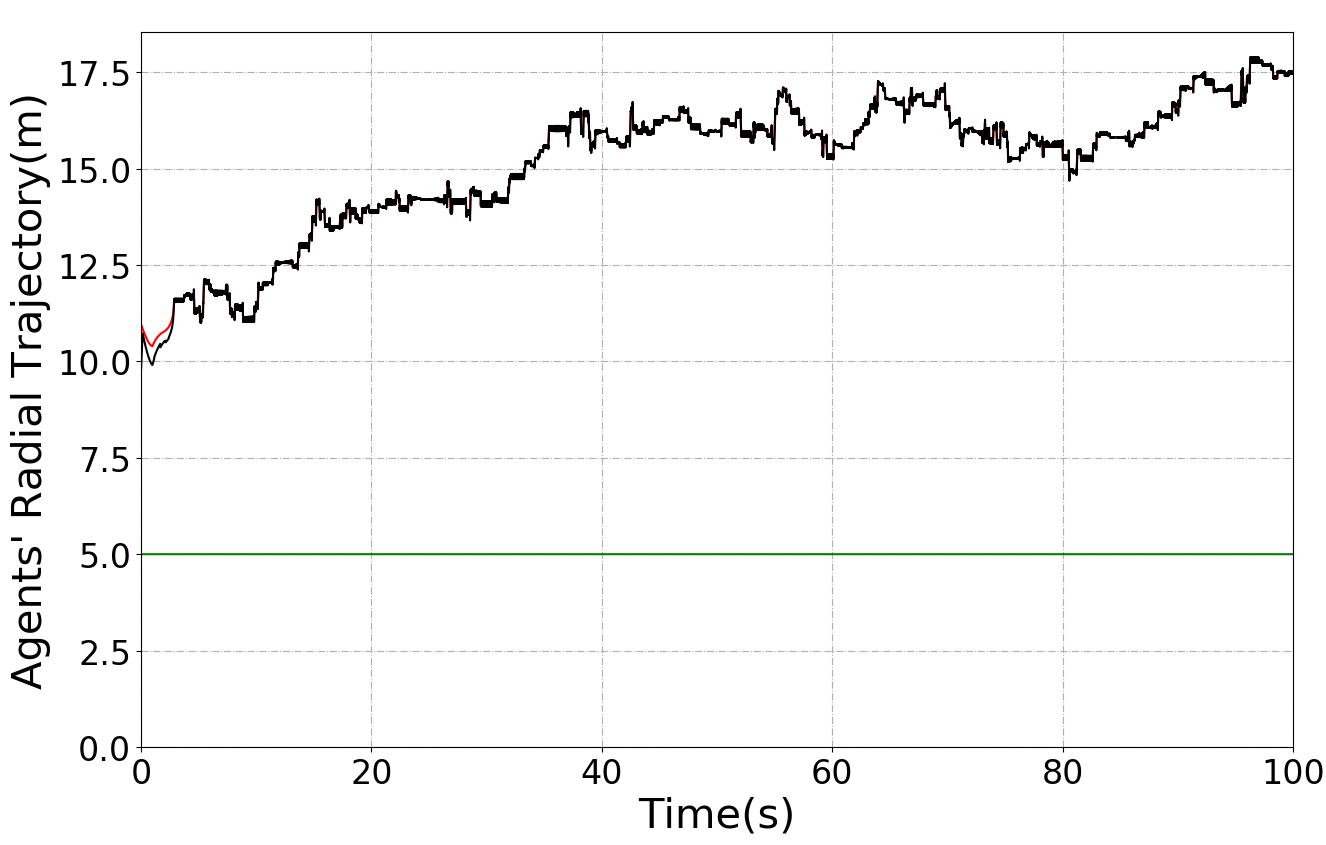}\label{fig:AT}}\\
\subfloat[ ]{\includegraphics[width= \linewidth]{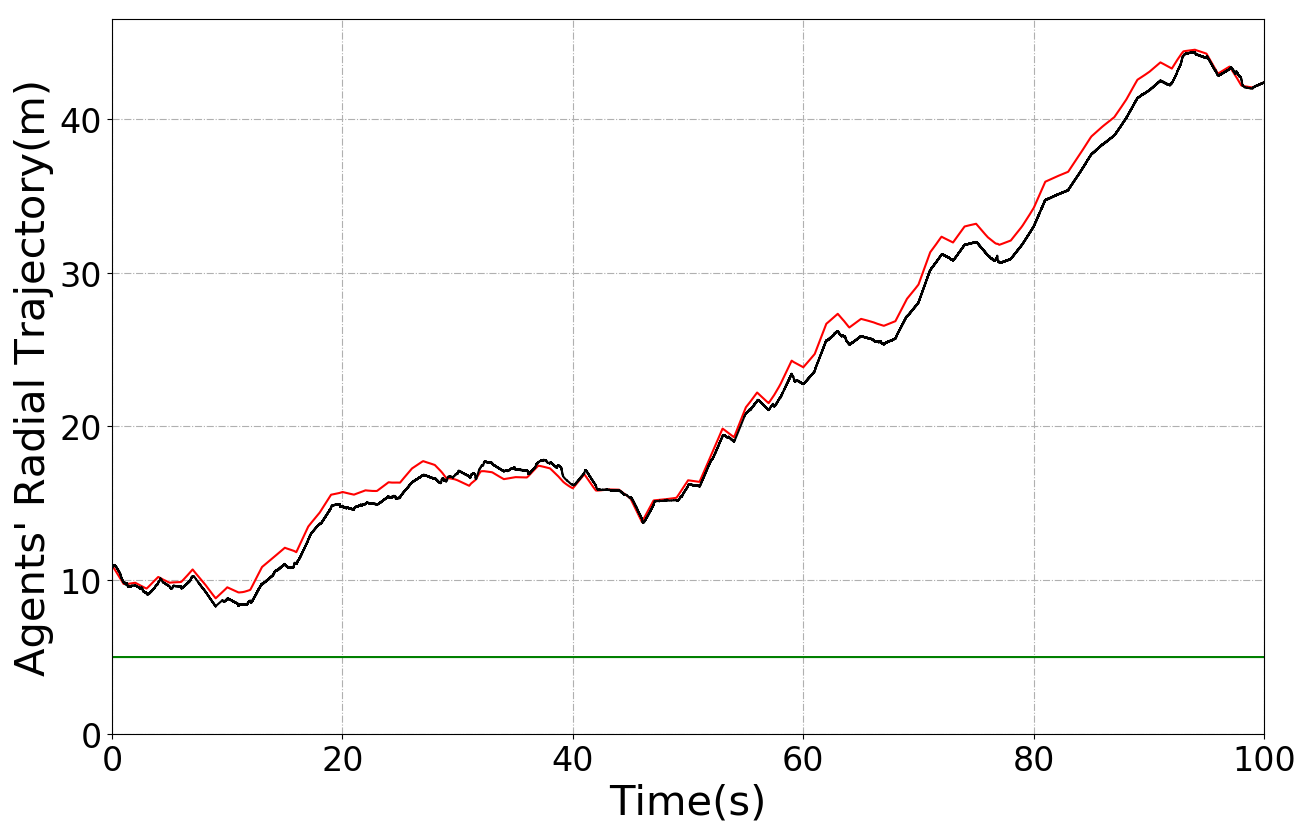}\label{fig:AQT}}
\caption{Radial coordinate of the target (red line) and of the herder (black line) driven by (a) the control law $v$ and (b) the classical Q-learning after 4000 trials. The green solid line shows the radius of the circular goal region.}
\end{figure}

Fig. \ref{fig:ATTVFB200} shows that the herder driven by CTQL is extremely effective in solving the problem, achieving  convergence of the target agent towards the goal region just after 1 trial; the transient being notably reduced as more learning trials are taken into account. 
On the contrary Fig.~\ref{fig:AT} and Fig.~\ref{fig:AQT} show that the problem cannot be solved if the control law \eqref{eq:4-9} or the classical Q-learning algorithm were to be used on their own to drive the herder agent.
(All the simulation parameters and a details necessary for the numerical implementation of CTQL can be found in \cite{DeLellis2019}.)

\begin{figure}[htbp]
\subfloat[ ]{\includegraphics[width=\linewidth]{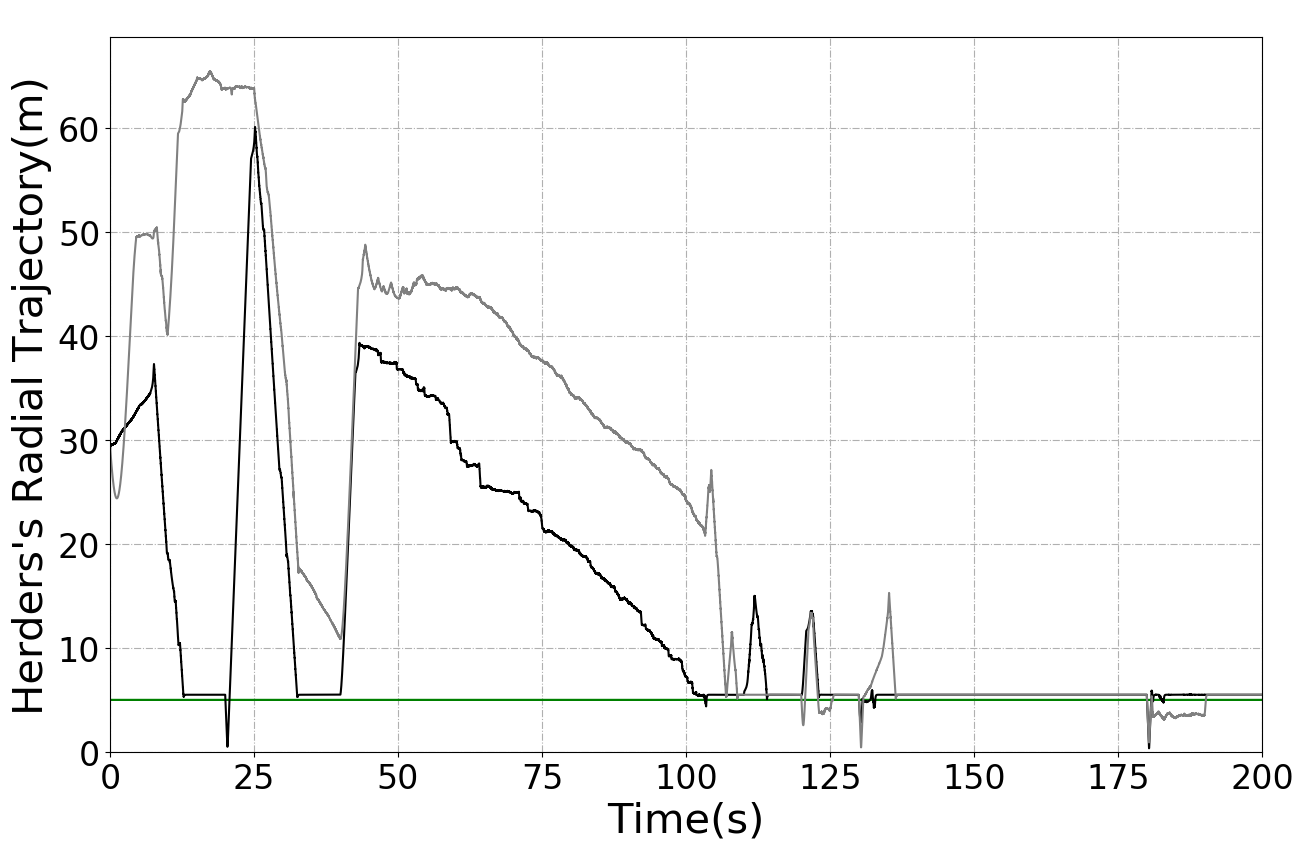}}\\
\subfloat[ ]{\includegraphics[width=\linewidth]{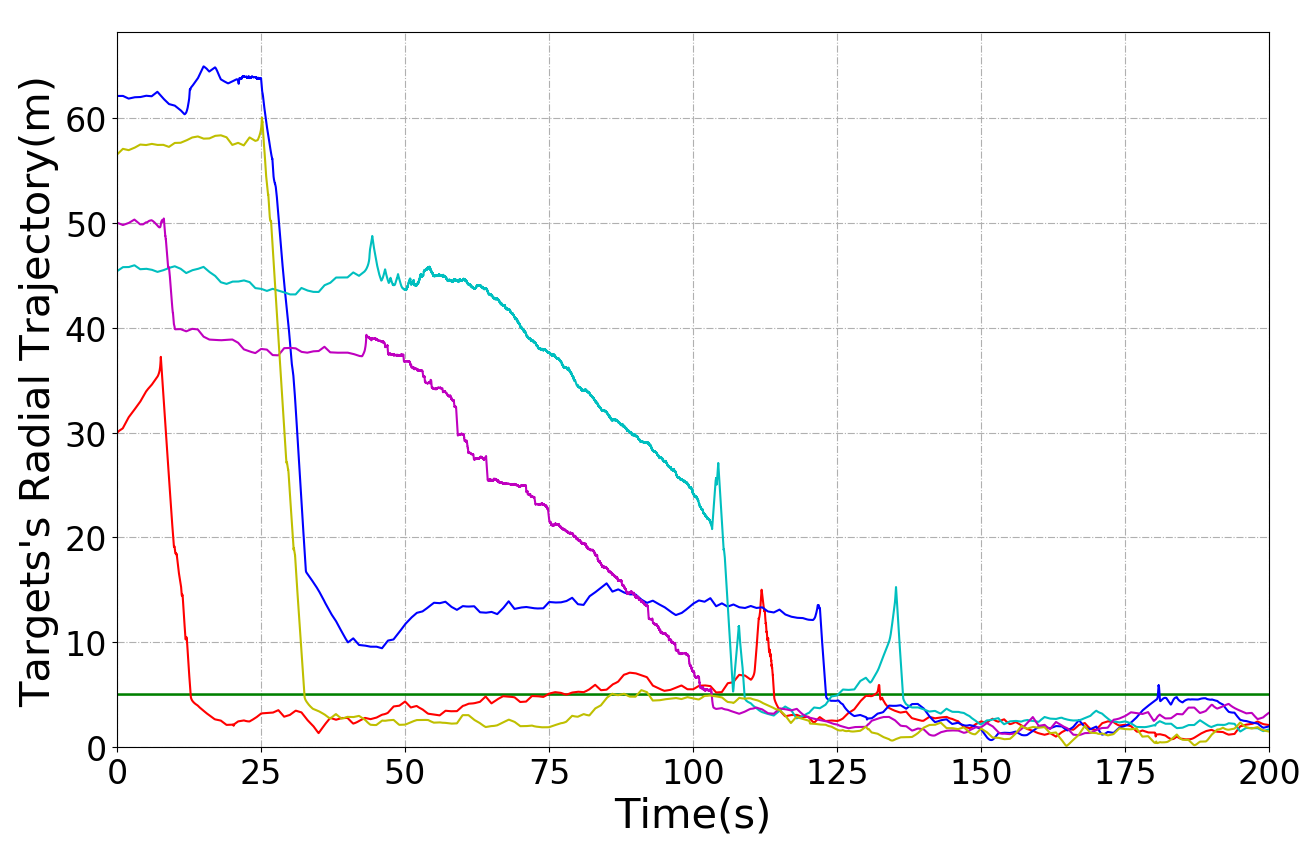}}
\caption{Radial coordinates of (a) $M=2$ herding agents and (b) $N=5$ target agents. As can be seen, after 1 trial, the herders driven by the CTQL successfully manage to collect all the targets inside the goal region (horizontal green line). Since the targets have some stochasticity they occasionally get out of the goal region but the herders are able to push them back inside the region.} 
\label{fig:ATTVFB5}
\end{figure}

As a further validation, the CTQL strategy was also used to solve the herding problem involving $M = 2$ herders controlling $N = 5$  targets (see \cite{DeLellis2019} for futher details). As shown in Fig. \ref{fig:ATTVFB5}, even in this harder case, herders controlled by CTQL successfully push all the targets inside the goal region and are able to recover any target that occasionally moves out because of random perturbations.

The application of CTQL to the herding problem proves that the combination of learning and feedback control can achieve ambitious control goals even in those cases where neither would work on its own. We envisage that a similar control tutored approach can be used to enhance the performance and convergence of other more sophisticated learning algorithms. 
Ongoing work is focused on refining this approach with the aim of obtaining a better understanding of its advantages and limitations for future applications \cite{DeLellis2019}.

\section*{Acknowledgements}
The authors wish to thank Pietro De Lellis at the University of Naples Federico II for his reading of an earlier version of this abstract and his insightful comments.

\bibliographystyle{ifacconf}
\bibliography{ExtendedAbstract}             

\end{document}